\documentclass{article}

\usepackage[preprint]{packages/corl_2026} 
\title{Reactive Task-Oriented Robot-Human Handovers via Generative Hypothesis Selection}

\author{
    Carmen Scheidemann* \quad Andreea Tulbure* \quad Pascal Burkhardt \quad Marco Hutter \\
    Robotic Systems Lab, ETH Zurich \\
    *equal contribution, corresponding: \texttt{carmensc@ethz.ch}
}

\usepackage[utf8]{inputenc}
\usepackage[OT1]{fontenc}

\usepackage{a4}

\usepackage{fancyhdr}

\usepackage{textcomp}\usepackage{gensymb}

\usepackage[english]{isodate}

\usepackage{multicol}

\usepackage{graphicx}

\usepackage{subfigure}

\usepackage{booktabs}
\usepackage{array}
\usepackage{multirow}
\usepackage{enumitem}

\usepackage{color}
\usepackage{colortbl}
\definecolor{black}{rgb}{0,0,0}
\definecolor{white}{rgb}{1,1,1}
\definecolor{darkred}{rgb}{0.5,0,0}
\definecolor{darkgreen}{rgb}{0,0.5,0}
\definecolor{darkblue}{rgb}{0,0,0.5}

\usepackage{amsmath}
\usepackage{amssymb}

\usepackage{nicefrac}

\usepackage{upgreek}

\usepackage{isomath}

\usepackage{rotating}

\usepackage{pdfpages}
\includepdfset{pages={-}, frame=true, pagecommand={\thispagestyle{fancy}}}

\usepackage{pgfgantt}

\PassOptionsToPackage{hyphens}{url}

\usepackage{cleveref}
\let\cref\Cref

\usepackage{tabularx}
\newcolumntype{Y}{>{\centering\arraybackslash}X}

\usepackage{makecell} 
\usepackage{siunitx}
\usepackage{multirow}
\usepackage{float}
\usepackage{fontawesome5} 
\usepackage{xcolor}       
\usepackage{booktabs}
\usepackage{ragged2e}
\usepackage{amssymb}
\usepackage{geometry}
\usepackage{caption}
\usepackage{siunitx}
\usepackage{threeparttable}
\definecolor{Gray}{gray}{0.95}

\usepackage{listings}
\definecolor{pygkw}{HTML}{008000}
\definecolor{pygstr}{HTML}{BA2121}
\definecolor{pygcom}{HTML}{3D7B7B}
\definecolor{pygnum}{HTML}{666666}
\begin{document}
\maketitle


\begin{abstract}
When humans hand each other objects, they incorporate both geometric and semantic information into this process. For example, passing a knife with the handle towards the recipient, rather than the blade, is both more ergonomic and safer. Recent state-of-the-art methods for task-oriented robot-human handovers have progressed from modeling object geometry to incorporating object affordances. However, they often forgo predicting the explicit, task-specific hand poses a human selects to utilize an object. Since many objects support multiple interaction modalities, e.g., a claw hammer used to strike or pull nails, this variability must be modeled to achieve robust task-oriented handovers. To tackle this, we propose a novel approach, GENESIS-Handover (GENErative HypotheSIS), which leverages VLM image generation to produce a variety of task-specific hand-object interaction hypotheses. These hypotheses are matched in real time to the observed human hand pose, enabling inference of the most suitable handover configuration. By leveraging VLMs as priors of plausible hand-object interactions, the method produces task-conditioned handover strategies for previously unseen object-task pairs. We evaluate the standalone interaction proposal module before deploying the full system on a mobile manipulator. In a user study with 12 participants across five task-object pairs, 83.3\% perceived our method to have better task understanding than the previous state of the art.
\end{abstract}

\section{Introduction}
\label{sec:intro}
Methods enabling robots to pass objects to humans have evolved from simple object transfer toward interactions that account for safety, coordination, and task requirements. 
In particular, during task-oriented handovers, objects are presented in configurations that facilitate their intended use. 
Imagine, for example, someone passing you a screwdriver: if you aim to use it immediately, you would probably prefer the handle to be pointing towards you as you receive it.
Previous strategies to enable this relied on deriving how to present an object from predicted graspable regions or affordances \cite{llmhandover,tulburehri26,contactHandover}, e.g. the screwdriver handle, or from geometric and ergonomic criteria \cite{meng2024,lagomarsino2023maximising}. While these strategies enable comfortable grasping, they do not explicitly model post-handover object use. This may result in configurations that are feasible but suboptimal for the intended task, causing the human to reorient the object in their hand, which requires time and attention.

Recent learning-based methods move beyond heuristic strategies by predicting grasp configurations from object properties \cite{meng2022} or synthesizing task-consistent handover configurations with generative models \cite{zhang2025generative}. While these approaches capture aspects of human tool-use behavior and enable more precise presentation, they typically predict only a single handover configuration. In contrast, humans often use multiple valid grasp strategies for the same object and task, and collapsing this variability to a single prediction can limit interaction flexibility and result in suboptimal handovers.


\begin{figure}[ht]
    \vspace{-10pt}
    \centering
    \includegraphics[width=0.95\linewidth]{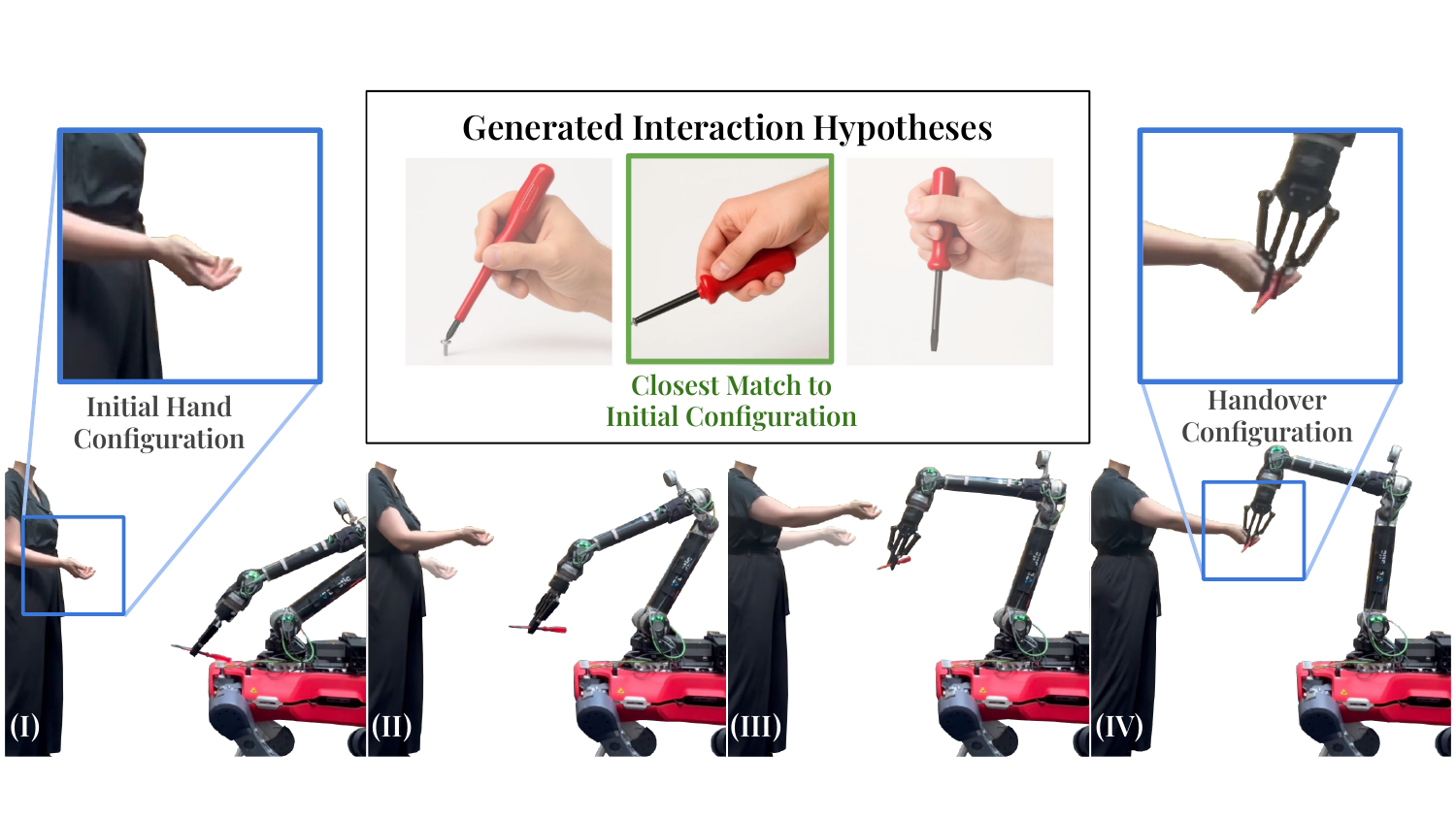}
    \vspace{-18pt}
    \caption{Example for the tool-task pair \emph{screwdriver, to screw}. Initially, GENESIS-Handover generates multiple hand-object interaction hypotheses via a VLM. At handover onset (I), the human hand pose is observed by the onboard camera, and each hypothesis is scored based on joint-level similarity to the observed pose. The closest hypothesis is selected for execution. During execution, the robot continuously adapts its target to the human hand (II and III, hand pose from previous frame blended in).  Finally, the handover occurs with the hand configuration matching the selected hypothesis (IV).}
    \label{fig:first-handover}
    \vspace{-15pt}
\end{figure}

In this work, we address these limitations by explicitly modeling intended object use through VLM image generation.
Rather than relying on heuristic presentation rules or predicting a single grasp configuration, we generate multiple diverse plausible hand-object interaction hypotheses, each representing a different hand-object configuration in which the object could be used after the handover. This enables the system to represent multiple plausible task-conditioned grasp configurations instead of collapsing interaction variability into a single estimate.
As depicted in Fig.~\ref{fig:first-handover}, at handover onset, the system observes the human hand and selects the interaction hypothesis that best matches the user’s hand configuration. During execution, the robot continuously tracks the human hand and adapts to its motion in real time. In this way, the method combines generative modeling of task-conditioned hand-object interactions with continuous alignment to the user’s hand pose. 

We evaluate the proposed approach through targeted data analysis and a user study with a legged manipulator, demonstrating that generative interaction hypotheses improve immediate usability, particularly for tool-oriented tasks, and better align with the user’s observed grasp configuration than heuristic presentation strategies \cite{tulburehri26}. 
The main contributions of this work are:
\begin{itemize}
    \item GENESIS-Handover: a framework that formulates task-oriented handovers as a multimodal hand-object interaction hypothesis selection problem, leveraging generative models. 
    \item Full system integration on a legged manipulator relying only on on-board sensing.
    \item A user study revealing improved subjective and objective handovers over a heuristic baseline for a variety of task-oriented handovers. We show a 2.67x increase in handovers requiring no human adjustment and an 83.3\% improvement in perceived task understanding.
\end{itemize}

\section{Related Work}
\label{chapter6-related-work}

Robotic handovers have long been studied as a core component of human-robot interaction. Foundational work shows that successful handovers depend not only on transfer stability but also on presenting objects in configurations that are comfortable and intuitive for the receiver \cite{Strabala2013Seamless, Cakmak2011Preferences}. 
Building on this, task-oriented handovers aim to present objects in ways that facilitate their intended use after transfer, requiring reasoning beyond geometric feasibility. 
Accordingly, we review prior work in two relevant directions: (I) object presentation strategies for task-oriented handovers, and (II) generative modeling approaches for human-object interaction.

\subsection{Object Presentation for Task-Oriented Handovers}
A common approach to object presentation in task-oriented robot-human handovers is to derive grasp configurations from predicted affordances or likely human grasp regions \cite{llmhandover,tulburehri26,contactHandover}, orienting the object to make these regions accessible. While such strategies improve accessibility and grasp comfort, they do not explicitly model post-handover object use.

Other approaches optimize geometric and ergonomic criteria: Chan et al. \cite{chan2020affordance} combine affordance reasoning with distance minimization, while Meng et al. \cite{meng2024} adapt handover poses based on reachability, safety, and comfort. Similarly, Lagomarsino et al. \cite{lagomarsino2023maximising} use reinforcement learning to optimize object orientation, distance, and motion from human-centric signals such as comfort and attention. Although these methods improve interaction quality, object presentation remains driven primarily by geometric or ergonomic considerations rather than intended object use.

Methods that explicitly account for intended use include affordance-aware optimization combining task context, object affordances, and human mobility constraints \cite{Ardon2020}, collaborative tool-use handovers based on functional object regions \cite{Meiying2022}, and learning-based grasp prediction from object properties \cite{meng2022}. However, these approaches typically predict a single hand-object configuration, limiting their ability to capture the variability of human grasping.

\subsection{Generative Modeling of Human-Object Interaction}
Generative modeling has emerged as a promising paradigm for capturing the diversity of human-object interactions. In computer vision and graphics, diffusion-based methods such as Text2HOI \cite{cha2024text2hoi} and DiffH2O \cite{christen2024diffh2o} generate plausible interactions conditioned on object geometry or textual descriptions, while generative grasp synthesis methods model distributions over feasible grasps \cite{barad2024graspldm}.
These ideas have also been applied to robotic manipulation and handovers. GRIM-TOG \cite{raj2026grim}, for example, models human grasp behavior to guide robot grasp synthesis, while \cite{zhang2025generative} generates task-consistent handover configurations by predicting human hand poses and object placements. Although such methods capture aspects of grasp diversity, they ultimately collapse this variability into a single predicted configuration. 

Overall, task-oriented robot-to-human handovers commonly determine object presentation through (I) heuristic or affordance-based rules, (II) geometric and ergonomic optimization, or (III) learned or generative prediction of a single grasp configuration. While these approaches improve accessibility and comfort, they typically approximate task-oriented handovers with a single grasp configuration, limiting their ability to model the multimodal structure of human grasp behavior. In contrast, we model object presentation through synthetic hand-object interaction hypotheses capturing different ways an object may be used, enabling handovers that better align with the user’s intended grasp. 


\begin{figure*}[h]
    \centering
    \vspace{-3pt}
    \includegraphics[width=0.95\linewidth]{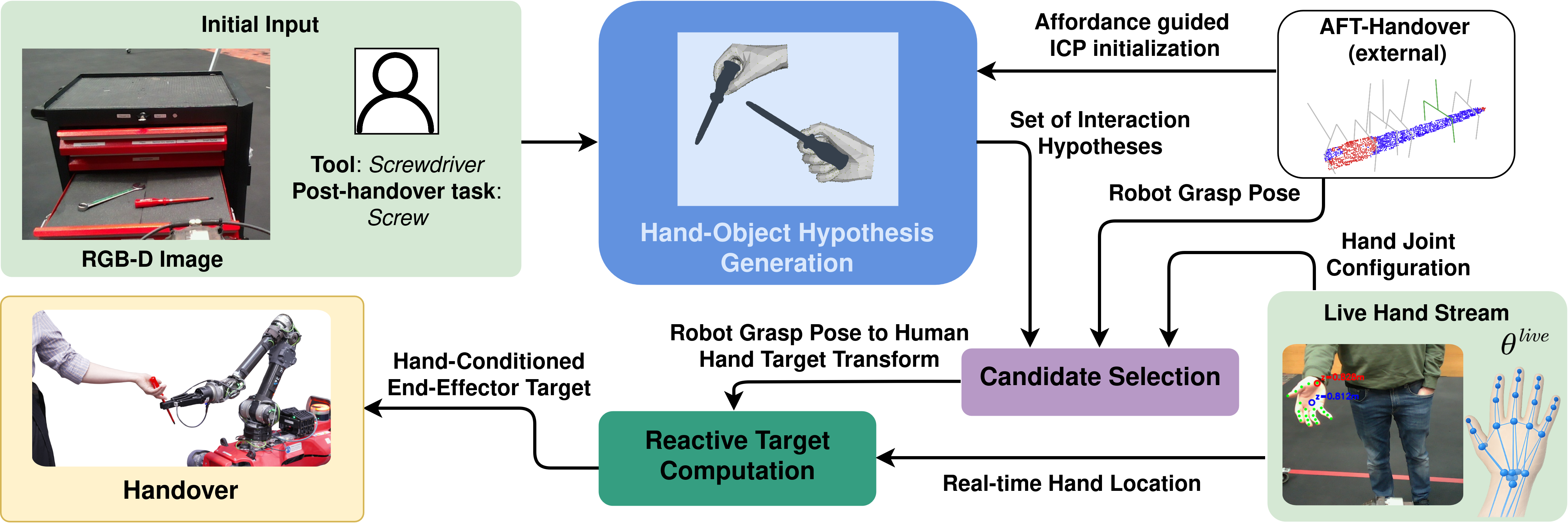}
    \caption{Overview of GENESIS-Handover. The system generates task-conditioned hand-object interaction hypotheses (blue), selects the best-matching hypothesis based on the observed human hand (purple), and computes a reactive end-effector target by anchoring the selected relation to the live hand pose (dark green). AFT-Handover provides affordance priors for alignment initialization and object placement in the gripper (white). All inputs to the system are shown in light green.
    }
    \vspace{-5pt}
    \label{fig:methodology_overview}
\end{figure*}

\section{Methodology}
\label{chapter6-method}

The GENESIS-Handover framework, as shown in Fig.~\ref{fig:methodology_overview}, decomposes reactive task-oriented handovers into three stages: Hand-Object-Hypothesis Generation, Candidate Selection, and Reactive Target Computation. 
The initial inputs to the system are an RGB-D image of the scene, the object name, and the post-handover task description (e.g., \emph{screwdriver, screw}). The text input could alternatively be generated by an additional orchestration language model, as in~\cite{zhang2025generative}. From the input, we generate a discrete set of task-conditioned hand-object interaction hypotheses, each defining a distinct hand-object transform converted into an end-effector target. At handover onset, the robot observes the receiver’s hand configuration and selects the hypothesis whose articulation most closely matches the observed user hand configuration. 
During execution, the transform of the selected hypothesis is anchored to the observed hand pose, producing a continuously updated end-effector target.

\begin{figure*}[h]
    \centering
    \vspace{-5pt}
\includegraphics[width=0.94\linewidth]{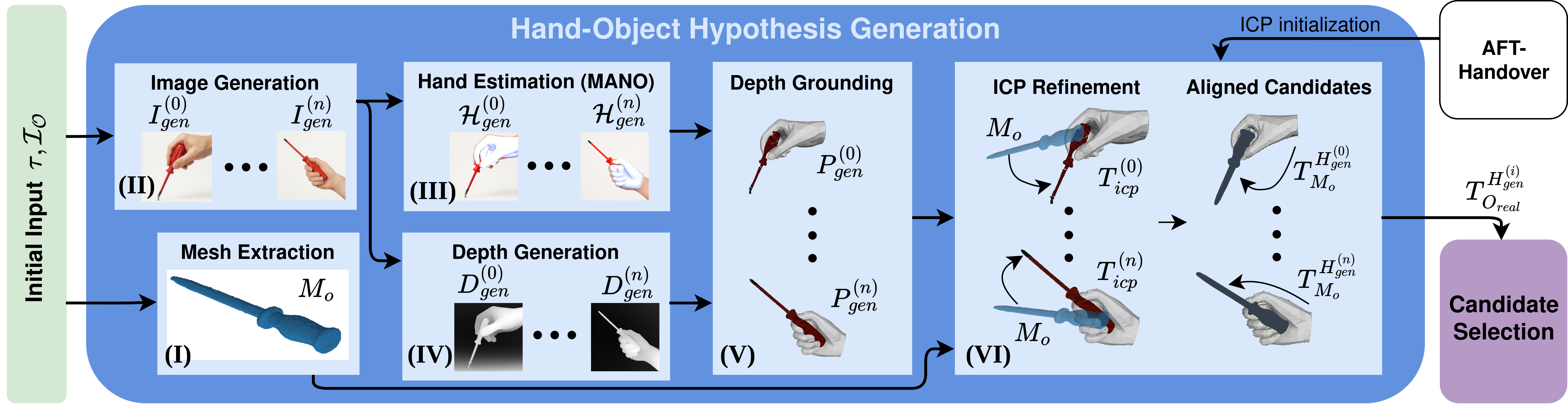}
    \caption{Overview of the \emph{Hand-Object Hypothesis Generation} block. From an RGB-D scene observation and task specification, the system generates multiple task-conditioned hand-object hypotheses and maps them into the real object frame: The system first reconstructs the object mesh (I), then generates multiple synthetic hand-object interaction hypothesis images (II), generates depth and hand pose estimates (III and IV), lifts these into metric 3D (V), and expresses each hypothesis as a relative transform between the human hand and the real object (VI).
    }
    \vspace{-10pt}
    \label{fig:methodology_hohg}
\end{figure*}

\subsection{Interaction Image Generation and Depth Grounding}
As visualized in Fig.~\ref{fig:methodology_hohg}, given a task description $\tau$ and an RGB-D image of the object $\mathcal{I_O}$, we generate $N$ synthetic interaction images (step II):
\vspace{-4pt}
\begin{equation}
I_{gen}^{(i)} = \mathrm{ImgGen}(\tau, \mathcal{I_O}),
\quad i = 1,\dots,N
\end{equation}
each representing different plausible grasp configurations for the task.
To obtain diverse synthetic images of the hand-object interaction, we employ a two-stage generative pipeline. First, we prompt an LLM to produce $N$ maximally diverse grasp-style descriptions for the given object-task pair. 
In the second stage, each grasp-style description is used to condition an image-generation model, instructing it to generate an image of a human hand holding the object and performing the specified task, preserving the exact object geometry and appearance from the real RGB image of the object. 

For each generated image $I_{gen}^{(i)}$, a hand mesh $\mathcal{H}_{gen}^{(i)}$ is estimated (step III) with $\mathcal{H}_{gen}^{(i)} = \mathrm{MANO}({\beta}^{(i)}, {\theta}^{(i)})$, using MANO shape $\beta$ and pose parameters $\theta$ \cite{mano2017}.
A corresponding depth map $D_{gen}^{(i)}$ is generated via monocular depth estimation~\cite{depthanythingv2} (step IV). To ground each depth estimate, we utilize $\mathcal{H}_{gen}^{(i)}$: The MANO mesh is aligned with the hand point cloud in the virtual camera frame (step V), enabling recovery of the global scale and expression of both the hand and the synthetic object point clouds $P^{(i)}_{gen}$ in a metrically consistent space.



\subsection{ICP Refinement and Candidate Alignment}
To bridge the gap between the synthetic and real images, for each hypothesis, we align the real object mesh $M_o$ (extracted in step I) with the corresponding synthetic object point cloud $P^{(i)}_{gen}$ using ICP (step VI). To support alignment initialization, the framework leverages task-specific affordance priors from AFT-Handover \cite{tulburehri26}, which provide a rough estimate of how the object should be oriented relative to the human hand.
This alignment yields transforms $T^{(i)}_{icp}$ that map $M_o$ into the synthetic camera frame as $T^{(i)}_{M_o}$. With both the synthetic hand and real object expressed in the same frame, we compute each relative transform: 
\vspace{-12pt}
\begin{equation}
T^{H_{gen}^{(i)}}_{M_o} = (T^{(i)}_{H_{gen}})^{-1} \cdot T^{(i)}_{M_o}
\end{equation}
Yielding a set of transforms representing all candidate hand configurations, mapped to the real scene. 

 
\subsection{Candidate Selection}
\label{sec:hand-similarity}
Before initializing the handover, the robot observes the configuration of the human hand using its on-board RGB-D camera and selects the most compatible hypothesis based on joint-level hand articulation similarity.
To ensure smooth, predictable robot motion, the closest hypothesis is kept static after this initial selection.
Let $\{\theta^{(i)}\}_{i=1}^{N}$ denote the MANO pose parameters of the generated hands and $\theta^{(\mathrm{live})}$ the MANO parameters estimated from the live observation. For each hand, we reconstruct the MANO joints and express them in a wrist-centered coordinate frame, yielding $\hat{\mathbf{J}}^{(i)} \in \mathbb{R}^{K \times 3}$ and $\hat{\mathbf{J}}^{(\mathrm{live})} \in \mathbb{R}^{K \times 3}$. The joint-configuration distance is then computed as:
\vspace{-5pt}
\begin{equation}
D^{(i)} = \sum_{k=1}^{K}
\left\|
\hat{\mathbf{J}}^{(\mathrm{live})}_k -
\hat{\mathbf{J}}^{(i)}_k
\right\|_2^2,
\quad i=1,\dots,N.
\end{equation}
The closest reference hand is selected as 
$D^{(sel)} = \arg\min_{i \in \{1,\dots,N\}} D^{(i)}$,
yielding the closest hypothesis, $T^{H_{gen}^{(sel)}}_{O_{real}}$, where ${O_{real}}$ is the root frame of the object. On the robot side, we assume an accurate pickup grasp and a static grasp during transfer and do not compensate for grasp slippage or execution grasp-pose errors.
The robot's grasping point on the object, $T^{O_{real}}_{ee}$, is generated by AFT-Handover.
From these, we can express the desired relation between the human hand and robot gripper as a static transformation:
\vspace{-12pt}
\begin{equation}
   T^{H_{gen}^{(sel)}}_{ee}=T^{H_{gen}^{(sel)}}_{O_{real}} \cdot T^{O_{real}}_{ee}
\end{equation}

\subsection{Reactive Target Computation}
Applying a hand pose detection model to the RGB-D stream from the robot’s on-board camera yields an updated human hand pose estimate at each time step, $T_{H_{live}}^W$. 
Assuming $T_{ee}^{H_{live}} \approx T^{H_{gen}^{(sel)}}_{ee}$, we continuously update the handover pose to adapt to the observed hand motion:
\vspace{-3pt}
\begin{equation}
T^{W^*}_{ee} = T_{H_{live}}^W \cdot    T^{H_{gen}^{(sel)}}_{ee}
\end{equation}
\vspace{-20pt}
\section{Evaluation}
\label{chapter6-evaluation}
In this section, we first analyze the generated grasp hypotheses, then present the core findings of the user study. For additional evaluations of GENESIS-Handover, including analysis of the hypothesis selection, as well as system reactivity, we kindly refer the reader to the appendix.
\subsection{Experimental Setup and Implementation Details}
We use an ANYmal robot with a top-mounted arm, equipped with an L515 Realsense RGB-D camera and a Robotiq 2f gripper. We replicate the experimental setup described in \cite{tulburehri26}, substituting the handover pose computation and execution with GENESIS-Handover. In summary, a behavior tree runs AFT-Handover \cite{tulburehri26} to compute the grasp affordances and select a grasp, executes object pickup, turns around to look for the human, approaches them once detected, starts hand detection once they are in the workspace, and executes the handover. Motion planning is done via the MPC-based framework of \cite{sleiman2021unified}.
The computations for real-time, reactive interaction run on the on-board NVIDIA Jetson Orin: WiLoR \cite{wilor2025} for hand tracking and YOLOv8 \cite{yolov8} for human detection. Motion planning and behavior tree execution also run onboard. Hypothesis generation is performed on an external workstation with one NVIDIA RTX 4090 and one RTX 3070 GPU. 
For image generation, we use OpenAI's GPT Image 1, while for the textual grasp-style descriptions, we use OpenAI's GPT-5-3. We use the original AFT-Handover models from \cite{tulburehri26}, SAM 3D \cite{sam3d} for mesh reconstruction, DepthAnythingv2 \cite{depthanythingv2} for depth estimation, WiLoR \cite{wilor2025} for synthetic-image hand mesh estimation, LangSAM \cite{LangSAM} for segmentation, and Open3D \cite{open3d} for ICP.


\subsection{Generative Grasp Quality and Coverage}
\label{sec:contactdb-analysis}
We first evaluate the generated grasp hypotheses independently of the downstream selection and execution components to verify that the generative model produces realistic and expressive human grasp configurations. As reference data, we use the ContactPose dataset~\cite{contactpose}, which contains multiple human grasp demonstrations per object and captures variability across users.
We select 12 representative objects, each featuring a task-specific use case, from the 50-object dataset. All results use $N=10$ generated hypotheses per object, resulting in 120 generated hand-object interactions. We define diversity as the global spread of grasp configurations, measured by the median distance between two grasps as defined in \ref{sec:hand-similarity}. 
Higher human grasp diversity (H. Div.) indicates multiple grasp modes and thus more multimodal grasping behavior.

\begin{table}[h]
\centering
\vspace{-12pt}
\caption{Grasp distribution analysis. Distances are symmetric surface distances (in mm). 
}
\label{tab:grasp_distribution}
\resizebox{1.0\columnwidth}{!}{
\begin{tabular}{lcccccccccccc}
\toprule
Metric
& Mug
& Pan
& Apple
& Hammer
& Knife
& Cellphone
& Cup
& Flashlight
& Scissors
& Toothbrush
& Teapot
& Wine Glass \\
\midrule

H. Div.
& 15.1
& 10.5
& 28.0
& 10.4
& 10.2
& 9.5
& 31.1
& 8.7
& 18.3
& 12.9
& 12.1
& 40.0 \\

Gen$\rightarrow$CP $\downarrow$
& 8.3
& 11.6
& 14.3
& 8.7
& 8.8
& 9.1
& 12.8
& 11.4
& 9.0
& 11.5
& 11.8
& 9.1 \\

CP$\rightarrow$Gen $\downarrow$
& 11.3
& 12.5
& 16.2
& 10.2
& 9.2
& 9.2
& 14.7
& 11.8
& 10.7
& 10.3
& 10.6
& 14.5 \\

\bottomrule
\end{tabular}
}
\vspace{-8pt}
\end{table}

To evaluate the quality of the generated grasp hypotheses, we analyze realism and coverage using symmetric nearest-neighbor distances between generated and ContactPose grasps. For each generated grasp, the generated-to-ContactPose ($Gen \rightarrow CP$) distance measures the distance to the closest human grasp. 
Conversely, the ContactPose-to-generated ($CP \rightarrow Gen$) distance measures, for each human grasp, the distance to the closest generated grasp. 
Hence, Low $Gen\rightarrow CP$ distances indicate high realism, while low $CP \rightarrow Gen$ distances indicate good coverage. 
As shown in Table~\ref{tab:grasp_distribution}, $Gen \rightarrow CP$ distances are consistently low across objects ($\leq 15 mm$), indicating that generated grasps are geometrically close to human demonstrations and therefore realistic. In contrast, $CP \rightarrow Gen$ distances vary across objects, revealing variability in coverage of the human grasp distribution.
Objects such as \emph{hammer} or \emph{knife} exhibit low distances in both directions, indicating both realism and good coverage. For objects such as \emph{apple}, \emph{cup}, and \emph{wine glass}, higher $CP \rightarrow Gen$ distances suggest that the generated hypotheses do not fully capture the variability of human grasps. This effect is largely due to the unconstrained nature of grasping for these objects, partially due to their rotational symmetry; humans can grasp the object at many different locations with similar functionality.
This analysis, together with a comparison to AFT-Handover~\cite{tulburehri26} in Appendix~\ref{ap:baseline}, shows that the generated grasp hypotheses are geometrically close to human demonstrations and capture distinct, precisely localized grasp modes rather than an averaged contact distribution, providing meaningful coverage particularly for objects with lower grasp diversity.

\subsection{System Evaluation and User Study}
\label{sec:user-study}
To evaluate how GENESIS-Handover performs in real-world interactions, we conducted a user study. We assess whether the proposed formulation improves alignment between the presented object orientation and the user’s observed grasp configuration and immediate task usability, compared to the heuristic object presentation baseline of AFT-Handover. To ensure comparable conditions, both methods share the same grasp detection, motion planning, and execution modules. Additionally, the number of hypotheses was limited to $N =2$ to maintain comparable computation times.
Participants interacted with a legged manipulator performing handovers across five object-task pairs: mug-drink, ladle-scoop, screwdriver-screw, hammer-hammer, and paintbrush-paint. In this within-subject study, 12 participants experienced both GENESIS-Handover and the baseline AFT-Handover, resulting in 60 handovers per method. In each handover run, the robot was initialized in front of the object. We ran the respective method's pipeline, after which the robot picked up the object, approached the participant, and executed the handover. Participants were instructed to perform the specified task immediately after receiving the object.

\subsubsection{Evaluation Metrics}
\label{sec:eval-metrics}
We assess handover quality using both subjective and objective measures. After each interaction, participants rated robot task understanding, immediate object usability, and the robot's chosen grasp suitability for the task-oriented handover on a Likert scale (1-5). Additionally, participants rated overall motion predictability, timing appropriateness, and safety and comfort, and reported binary method preference as well as main perceived system limitations.

To complement subjective ratings, we classify each handover by the amount of user adjustment required before task execution into four categories: \textit{Perfect:} the object could be used without any adjustment; \textit{Small adjustment}: requiring minor in-hand or hand pose changes; \textit{Big adjustment:} requiring grasp reconfiguration or noticeable arm movement; \textit{Failure:}, where the object cannot be used without breaking the interaction, e.g., requiring a two-handed regrasp, major repositioning, or resulting in object drop or unusable orientation. Classification was performed through visual inspection of recorded interaction videos without knowledge of the method.
For an extended analysis of the user study, we kindly refer the reader to the appendix.

\subsubsection{Comparison to State-of-the-Art}
Comparing GENESIS to AFT-Handover, no statistically significant differences were observed for motion predictability, timing appropriateness, grasp suitability, or safety and comfort. Since both methods share identical grasp detection and motion planning modules, this indicates comparable execution-level behavior, allowing us to attribute observed differences to object presentation alone.

\textbf{Task Understanding}
In comparative judgments, 10 out of 12 participants ($83\%$) perceived GENESIS-Handover as presenting objects in a manner more consistent with the specified task (binomial test, $p=0.039$). While object-level Likert ratings show no statistically significant differences, moderate effect sizes appear for tool-based tasks such as hammering and painting. 

\textbf{Immediate Usability}
Improvements in immediate usability are most pronounced for tool-based tasks. A statistically significant improvement is observed for paintbrush ($p=0.039$, $d=0.68$) and a moderate effect for hammer ($p=0.071$, $d=0.58$). No meaningful differences are observed for mug, ladle, or screwdriver. These results suggest that modeling multiple task-conditioned interaction hypotheses is particularly beneficial for tool-oriented tasks with less diffuse human grasp variability. 

\textbf{Overall Preference} In total, 8 of 12 participants reported an overall preference for the proposed method over baseline. Additionally, participants who perceived GENESIS-Handover as having better task understanding preferred it in $80\%$ of cases. 
This corresponds to a moderate effect size (Cramér’s $V=0.40$), suggesting a practically meaningful relationship. However, potentially due to the limited number of user study participants, the association does not reach statistical significance ($p=0.171$).
This indicates that perceived task understanding contributes to preference, particularly when accompanied by improved usability.

\begin{figure}
    \centering
    \vspace{-7pt}
     \includegraphics[width=0.9\linewidth]{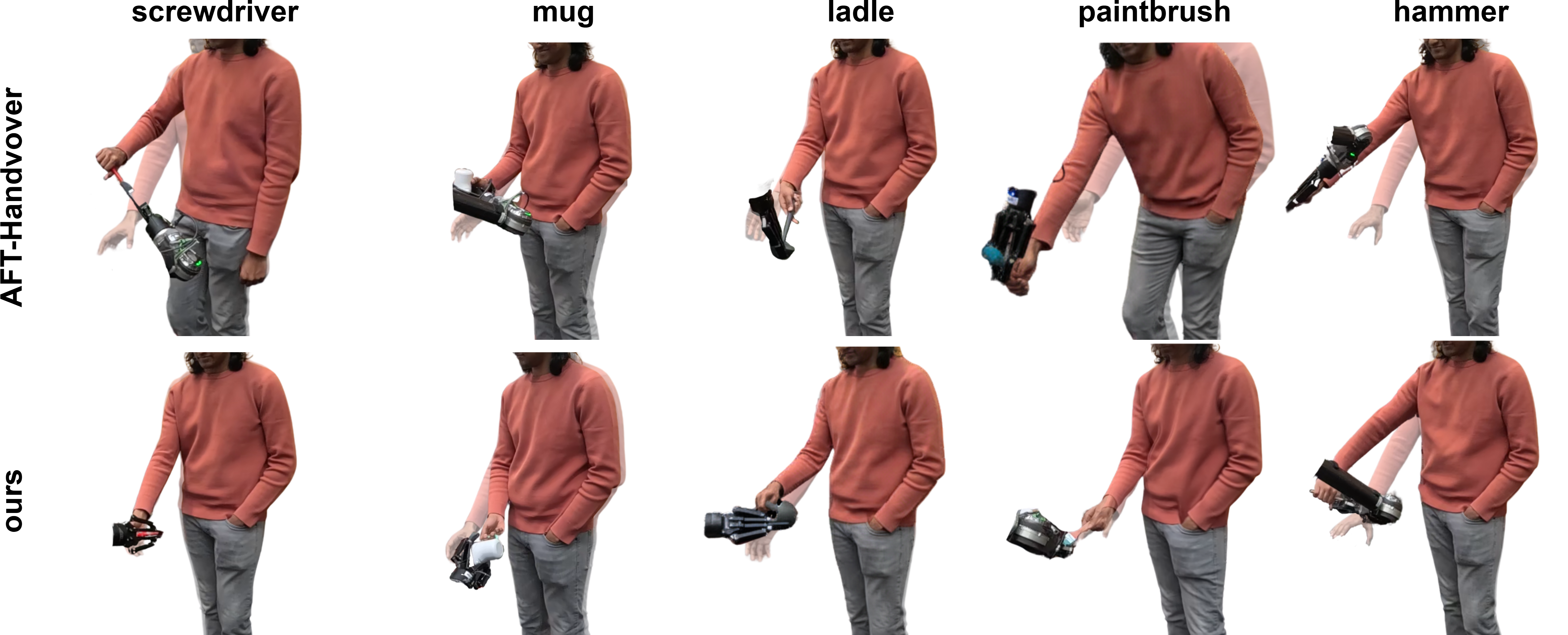}
     \vspace{-3pt}
     \caption{Example images comparing hand alignment between AFT-Handover and the proposed method for all handovers of one user study participant. GENESIS-Handover more closely matches the human hand configuration, requiring fewer adjustments to grasp the object than AFT-Handover. The initial handover pose is shown in transparent.}
     \label{fig:aft-ig-comparison-handover-loc}
     \vspace{-12pt}
\end{figure}

\textbf{Objective Handover Quality}
Consistent with subjective ratings, GENESIS-Handover improves objective handover quality, more than doubling perfect handovers (6 vs. 16) and substantially reducing those requiring big adjustments (21 vs. 11), as shown in Table~\ref{tab:handover-categ-comparison}. These results indicate that the proposed hypothesis-based formulation more frequently produces object presentations aligned with the user’s intended grasp, reducing the need for corrective motions prior to task execution.
Qualitative examples in Fig.~\ref{fig:aft-ig-comparison-handover-loc} further illustrate these differences, with GENESIS-Handover adapting more closely to the observed hand configuration.

\begin{table}[h]
\centering
\vspace{-15pt}
\caption{Handover distribution for GENESIS-Handover and AFT-Handover for the 60 user study handovers each.}
\vspace{-5pt}
\label{tab:handover-categ-comparison}
\resizebox{0.5\textwidth}{!}{
\begin{tabular}{lcccc}
\toprule
  \textbf{Method} &   Perfect&   Small &   Big &   Fail\\
\midrule
  GENESIS-Handover &   16 &   30 &   11 &   3\\
  AFT-Handover &   6 &   28 &   21 &   5\\
\bottomrule
\end{tabular}
}
\vspace{-5pt}
\end{table}

Overall, we observe a greater effect in the improvement of the objective handover quality than in the user preference. This difference could be explained by humans being subconsciously more ready to adapt to robots in short-term trials, and is potentially additionally influenced by novelty effects. However, this inefficiency could become more evident in the long term, especially considering repetitive collaboration tasks. 

\subsection{Hypothesis Selection Analysis}
\label{sec:appendix-hypothesis-sel}
We next evaluate the proposed hypothesis selection mechanism to understand when early hand observations are sufficient to select a compatible grasp hypothesis and how this relates to the system-level performance observed in the previous section. We analyze selection performance and consistency with the final grasp. Further analysis of the hypothesis structure can be found in \cref{ap:hyp_structure}. All the following analysis is performed on recorded interaction data from the user study using a fixed set of $N = 10$ generated hypotheses per object. Beyond the 60 handovers collected in the main study, we include an additional 25 handovers conducted under identical conditions. As these participants were not exposed to the baseline method, their data are considered only for system-level evaluation and are excluded from direct comparisons between methods.

\begin{table}[h]
\centering
\vspace{-12pt}
\caption{Per-object performance of hypothesis selection. 
\textit{$\Delta$ Rand.} denotes the relative reduction in average per-joint error compared to the mean random baseline. 
\textit{Oracle Gap} indicates the fraction of the random-to-oracle gap recovered. 
\textit{Better Rand.} reports the percentage of cases where the selected hypothesis achieves a lower error than the random baseline. 
\textit{Oracle Match} indicates the percentage of cases where the selected hypothesis coincides with the oracle.}
\vspace{2pt}
\label{tab:per_object_results}
\resizebox{0.65\linewidth}{!}{
\begin{tabular}{lccccc}
\toprule
\scriptsize \textbf{Object} & \scriptsize  n & \scriptsize  $\boldsymbol{\Delta}$ Rand. $\uparrow$ & \scriptsize  Oracle Gap $\uparrow$ & \scriptsize  Better Rand. $\uparrow$ & \scriptsize  Oracle Match $\uparrow$ \\
\midrule
\scriptsize  Hammer            & 8  & 16.6\% & 77.8\% & 100.0\% & 75.0\% \\
\scriptsize  Ladle             & 9  & 12.4\% & 62.0\% & 75.0\%  & 62.5\% \\
\scriptsize  Mug               & 7  & 19.3\% & 35.3\% & 80.0\%  & 0.0\%  \\
\scriptsize  Paintbrush        & 7  & 14.9\% & 54.7\% & 85.7\%  & 42.9\% \\
\scriptsize  Screwdriver   & 6  & 19.2\% & 49.2\% & 75.0\%  & 50.0\% \\
\midrule
\scriptsize  \textbf{Overall}  & 37 & \textbf{16.2\%} & \textbf{56.9\%} & \textbf{83.3\%} & \textbf{47.5\%} \\
\bottomrule
\end{tabular}
}
\vspace{-12pt}
\end{table}

\subsubsection{Selection Performance}
To evaluate the hypothesis selection, we compare the selected hypothesis against two baselines: random sampling from the generated set and an oracle that selects the hypothesis closest to the final observed grasp. Distances are computed as average joint deviation in hand configuration space.
As shown in Table \ref{tab:per_object_results}, the selected hypothesis reduces the average joint error by $16.2\%$ relative to random selection.
Relative to the oracle, which represents the best performance achievable within the hypothesis set, the selected hypothesis recovers $56.9\%$ of the oracle gap over all handovers, with a median of $94.1\%$, suggesting that selection often identifies compatible, though not always optimal, grasps. 
It outperforms random selection in $83.3\%$ of cases. 
To verify that the oracle corresponds to a meaningful optimal hypothesis, we compare its distance to the final observed grasp with the significant hand-configuration change threshold ($0.027$). The oracle achieves an average per-joint deviation of $0.0147$, validating it as a meaningful baseline.

Performance varies across object categories: For structured tools such as hammers, results are strong. However, for mugs, oracle-gap recovery is lower ($35.3\%$) with no oracle matches, despite a $19.3\%$ improvement over random. Thus, a substantial portion of the achievable performance within the hypothesis set can be recovered, despite not consistently recovering the fully optimal hypothesis.

\subsubsection{Consistency Between Early and Final Hand Configurations}
To assess whether the initial hand configuration reliably signals grasp intent, we compare the hand configuration at handover onset with the final observed grasp configuration. We present the full results for the interactions where the hand is visible to the robot at handover (37 of 85) in \cref{ap:handover_behav}.
Although measurable differences exist between initial and final hand configurations, these are primarily due to grasp closure rather than changes in grasp strategy. 
When comparing the initial observation to the last pre-contact configuration, the deviation is small, indicating that the underlying grasp structure remains largely stable throughout the approach phase.
These results indicate that early hand observations generally provide a stable representation of the intended grasp strategy.

\section{Limitations} \label{sec:limitations}

Several limitations arise from the proposed system and the experimental setup, primarily affecting execution quality. 
Onboard perception is constrained by the field of view of the robot-mounted camera, leading to frequent occlusions of the human hand during the later stages of the handover. This reduces the availability and reliability of hand pose observations and, together with motion planning constraints, limits reactivity, resulting in slower or less predictable adjustments during interaction. Accordingly, participants identified slow speed and reactivity ($76\%$) as a main limitation.
Participants also frequently reported uncertainty about the timing of object release ($76\%$), stemming from the absence of an explicit release strategy or feedback signal during transfer. Additionally, in some cases, the method resulted in uncomfortable arm motions ($23\%$) or approach trajectories ($5\%$).

Residual errors in object presentation are not explicitly compensated for in the current system: The approach assumes no object slippage or grasp-pose inaccuracies, which can lead to small deviations in the presented object orientation. 
While these errors did not lead to handover failures, they resulted in slightly misaligned object presentations that required small user adjustments. Similarly, the current approach does not explicitly enforce implicit physical constraints such as gravity alignment. This was especially evident in the mug task, where one participant reported that the upside-down presentation, despite preserving handle accessibility, conflicted with intended use and was perceived as a lack of task understanding (reported as a system limitation in a total of $5\%$ of handovers). 


\section{Conclusion}
\label{sec:conclusion}

In this work, we presented GENESIS-Handover, a generative, hypothesis-based framework for reactive task-oriented robot-to-human handovers. The approach frames handover generation as a two-stage process: first, leveraging generative models to produce a discrete set of task-conditioned hand–object interaction hypotheses, then selecting and reactively executing the hypothesis that best matches the observed human hand configuration. By grounding synthetic interaction hypotheses in the real scene, the framework enables adaptive object presentation through online selection and reactive execution of task-conditioned interaction hypotheses.

Our evaluation shows that the generated hypotheses are both realistic and sufficiently expressive to capture a range of plausible human grasp strategies. Quantitative comparisons against ContactPose indicate that the generated grasps remain geometrically close to human demonstrations while covering the human grasp distribution across diverse object classes. In real-world experiments with a legged manipulator, the proposed approach improves task-oriented object presentation over a heuristic baseline, particularly for tasks requiring precise object orientation. It more than doubles perfect handovers and reduces the need for large user adjustments, while participants frequently perceived the object presentation strategy as more consistent with the specified task.



\clearpage
\acknowledgments{The authors would like to thank Olga Vysotska and Andrei Cramariuc, as well as the assigned reviewers, for giving much useful feedback and improvement suggestions throughout the writing process. Additionally, we would like to thank all the participants of the user study for taking the time to make our evaluations possible.}

\bibliography{bibliography/references.bib}

\begin{thebibliography}{25}
\providecommand{\natexlab}[1]{#1}
\providecommand{\url}[1]{\texttt{#1}}
\expandafter\ifx\csname urlstyle\endcsname\relax
  \providecommand{\doi}[1]{doi: #1}\else
  \providecommand{\doi}{doi: \begingroup \urlstyle{rm}\Url}\fi

\bibitem[Tulbure et~al.(2025)Tulbure, Zurbrügg, Grigat, and Hutter]{llmhandover}
A.~Tulbure, R.~Zurbrügg, T.~Grigat, and M.~Hutter.
\newblock Llm-handover: Exploiting llms for task-oriented robot-human handovers.
\newblock \emph{IEEE Robotics and Automation Letters}, pages 1--8, 2025.

\bibitem[Tulbure et~al.(2026)Tulbure, Scheidemann, Steiner, and Hutter]{tulburehri26}
A.~Tulbure, C.~Scheidemann, E.~Steiner, and M.~Hutter.
\newblock Task-oriented robot-human handovers on legged manipulators.
\newblock \emph{Proceedings of the 2026 ACM/IEEE International Conference on Human-Robot Interaction}, 2026.

\bibitem[Wang et~al.(2024)Wang, Liu, Ouporov, and Song]{contactHandover}
Z.~Wang, Z.~Liu, N.~Ouporov, and S.~Song.
\newblock Contacthandover: Contact-guided robot-to-human object handover.
\newblock In \emph{IEEE/RSJ International Conference on Intelligent Robots and Systems (IROS)}, 2024.

\bibitem[Meng et~al.(2024)Meng, Zhang, Zhao, and Lam]{meng2024}
C.~Meng, T.~Zhang, D.~Zhao, and T.~L. Lam.
\newblock Fast and comfortable robot-to-human handover for mobile cooperation robot system.
\newblock \emph{Cyborg and Bionic Systems}, 5:\penalty0 0120, 2024.

\bibitem[Lagomarsino et~al.(2023)Lagomarsino, Lorenzini, Constable, De~Momi, Becchio, and Ajoudani]{lagomarsino2023maximising}
M.~Lagomarsino, M.~Lorenzini, M.~D. Constable, E.~De~Momi, C.~Becchio, and A.~Ajoudani.
\newblock Maximising coefficiency of human-robot handovers through reinforcement learning.
\newblock \emph{IEEE Robotics and Automation Letters}, 8\penalty0 (8):\penalty0 4378--4385, 2023.

\bibitem[Meng et~al.(2022)Meng, Zhang, and Lam]{meng2022}
C.~Meng, T.~Zhang, and T.~L. Lam.
\newblock Fast and comfortable interactive robot-to-human object handover.
\newblock In \emph{2022 IEEE/RSJ International Conference on Intelligent Robots and Systems}, 2022.

\bibitem[Zhang et~al.(2025)Zhang, Dhafer, Hao, and Dong]{zhang2025generative}
H.~Zhang, A.~Dhafer, Z.~D. Hao, and H.~Dong.
\newblock A generative system for robot-to-human handovers: from intent inference to spatial configuration imagery.
\newblock \emph{arXiv preprint arXiv:2503.03579}, 2025.

\bibitem[Strabala et~al.(2013)Strabala, Lee, Dragan, Forlizzi, Srinivasa, Cakmak, and Micelli]{Strabala2013Seamless}
K.~Strabala, M.~K. Lee, A.~Dragan, J.~Forlizzi, S.~S. Srinivasa, M.~Cakmak, and V.~Micelli.
\newblock Toward seamless human-robot handovers.
\newblock \emph{J. Hum.-Robot Interact.}, 2013.

\bibitem[Cakmak et~al.(2011)Cakmak, Srinivasa, Lee, Forlizzi, and Kiesler]{Cakmak2011Preferences}
M.~Cakmak, S.~S. Srinivasa, M.~K. Lee, J.~Forlizzi, and S.~Kiesler.
\newblock Human preferences for robot-human hand-over configurations.
\newblock In \emph{2011 IEEE/RSJ International Conference on Intelligent Robots and Systems}, pages 1986--1993, 2011.

\bibitem[Chan et~al.(2020)Chan, Pan, Croft, and Inaba]{chan2020affordance}
W.~P. Chan, M.~K. Pan, E.~A. Croft, and M.~Inaba.
\newblock An affordance and distance minimization based method for computing object orientations for robot human handovers.
\newblock \emph{Int. Journal of Social Robotics}, 2020.

\bibitem[Ard{\'o}n et~al.(2021)Ard{\'o}n, Cabrera, Pairet, Petrick, Ramamoorthy, Lohan, and Cakmak]{Ardon2020}
P.~Ard{\'o}n, M.~E. Cabrera, E.~Pairet, R.~P. Petrick, S.~Ramamoorthy, K.~S. Lohan, and M.~Cakmak.
\newblock Affordance-aware handovers with human arm mobility constraints.
\newblock \emph{IROS}, 2021.

\bibitem[Qin et~al.(2022)Qin, Brawer, and Scassellati]{Meiying2022}
M.~Qin, J.~Brawer, and B.~Scassellati.
\newblock Task-oriented robot-to-human handovers in collaborative tool-use tasks.
\newblock In \emph{IEEE International Conference on Robot and Human Interactive Communication}, 2022.

\bibitem[Cha et~al.(2024)Cha, Kim, Yoon, and Baek]{cha2024text2hoi}
J.~Cha, J.~Kim, J.~S. Yoon, and S.~Baek.
\newblock Text2hoi: Text-guided 3d motion generation for hand-object interaction.
\newblock In \emph{Proceedings of the IEEE/CVF Conference on Computer Vision and Pattern Recognition}, pages 1577--1585, 2024.

\bibitem[Christen et~al.(2024)Christen, Hampali, Sener, Remelli, Hodan, Sauser, Ma, and Tekin]{christen2024diffh2o}
S.~Christen, S.~Hampali, F.~Sener, E.~Remelli, T.~Hodan, E.~Sauser, S.~Ma, and B.~Tekin.
\newblock Diffh2o: Diffusion-based synthesis of hand-object interactions from textual descriptions.
\newblock In \emph{SIGGRAPH Asia 2024 Conference Papers}, pages 1--11, 2024.

\bibitem[Barad et~al.(2024)Barad, Orsula, Richard, Dentler, Olivares-Mendez, and Martinez]{barad2024graspldm}
K.~R. Barad, A.~Orsula, A.~Richard, J.~Dentler, M.~A. Olivares-Mendez, and C.~Martinez.
\newblock Graspldm: Generative 6-dof grasp synthesis using latent diffusion models.
\newblock \emph{IEEE Access}, 12:\penalty0 164621--164633, 2024.

\bibitem[Raj et~al.(2026)Raj, Kumar, Shukla, Melnik, Beetz, Nandi, et~al.]{raj2026grim}
A.~Raj, N.~Kumar, P.~Shukla, A.~Melnik, M.~Beetz, G.~C. Nandi, et~al.
\newblock Grim: Task-oriented grasping with conditioning on generative examples.
\newblock In \emph{Proceedings of the AAAI Conference on Artificial Intelligence}, volume~40, pages 18118--18125, 2026.

\bibitem[Romero et~al.(2017)Romero, Tzionas, and Black]{mano2017}
J.~Romero, D.~Tzionas, and M.~J. Black.
\newblock Embodied hands: modeling and capturing hands and bodies together.
\newblock \emph{ACM Trans. Graph.}, 36\penalty0 (6), Nov. 2017.
\newblock ISSN 0730-0301.

\bibitem[Yang et~al.(2024)Yang, Kang, Huang, Zhao, Xu, Feng, and Zhao]{depthanythingv2}
L.~Yang, B.~Kang, Z.~Huang, Z.~Zhao, X.~Xu, J.~Feng, and H.~Zhao.
\newblock Depth anything v2.
\newblock \emph{Advances in Neural Information Processing Systems}, 37:\penalty0 21875--21911, 2024.

\bibitem[Sleiman et~al.(2021)Sleiman, Farshidian, Minniti, and Hutter]{sleiman2021unified}
J.-P. Sleiman, F.~Farshidian, M.~V. Minniti, and M.~Hutter.
\newblock A unified mpc framework for whole-body dynamic locomotion and manipulation.
\newblock \emph{IEEE Robotics and Automation Letters}, 6\penalty0 (3), 2021.

\bibitem[Potamias et~al.(2025)Potamias, Zhang, Deng, and Zafeiriou]{wilor2025}
R.~A. Potamias, J.~Zhang, J.~Deng, and S.~Zafeiriou.
\newblock Wilor: End-to-end 3d hand localization and reconstruction in-the-wild.
\newblock In \emph{Proceedings of the Computer Vision and Pattern Recognition Conference}, pages 12242--12254, 2025.

\bibitem[Varghese and M.(2024)]{yolov8}
R.~Varghese and S.~M.
\newblock Yolov8: A novel object detection algorithm with enhanced performance and robustness.
\newblock In \emph{2024 International Conference on Advances in Data Engineering and Intelligent Computing Systems (ADICS)}, pages 1--6, 2024.

\bibitem[Chen et~al.(2025)Chen, Chu, Gleize, Liang, Sax, Tang, Wang, Guo, Hardin, Li, et~al.]{sam3d}
X.~Chen, F.-J. Chu, P.~Gleize, K.~J. Liang, A.~Sax, H.~Tang, W.~Wang, M.~Guo, T.~Hardin, X.~Li, et~al.
\newblock Sam 3d: 3dfy anything in images.
\newblock \emph{arXiv preprint arXiv:2511.16624}, 2025.

\bibitem[Lan()]{LangSAM}
Langsam.
\newblock \url{https://github.com/luca-medeiros/lang-segment-anything}.
\newblock Accessed: 2025-03-10.

\bibitem[Zhou et~al.(2018)Zhou, Park, and Koltun]{open3d}
Q.-Y. Zhou, J.~Park, and V.~Koltun.
\newblock {Open3D}: {A} modern library for {3D} data processing.
\newblock \emph{arXiv:1801.09847}, 2018.

\bibitem[Brahmbhatt et~al.(2020)Brahmbhatt, Tang, Twigg, Kemp, and Hays]{contactpose}
S.~Brahmbhatt, C.~Tang, C.~D. Twigg, C.~C. Kemp, and J.~Hays.
\newblock {ContactPose}: A dataset of grasps with object contact and hand pose.
\newblock In \emph{The European Conference on Computer Vision (ECCV)}, August 2020.

\end{thebibliography}

\clearpage 
\appendix

\counterwithin{figure}{section}
\counterwithin{table}{section}
\counterwithin{equation}{section}
\phantomsection

\begin{center}
    {\large \bf Reactive Task-Oriented Robot-Human Handovers \\ via Generative Hypothesis Selection} \\[1.0ex]
    {\large Supplementary Material} \par
    \vspace{1.0em}
\end{center}

\section{Auxiliary Evaluations}
\subsection{Full Analysis of Generative Grasp Quality and Coverage}
To supplement the condensed analysis presented in the main paper, we here present our full analysis of the synthetic grasps from GENESIS-Handover.
We consider their realism, coverage, and diversity, and briefly compare against AFT-Handover as a baseline for grasp region prediction. 


For this analysis, we select 14 representative objects out of the 50 everyday objects in the ContactPose dataset~\cite{contactpose}. As in the main paper, all results in this analysis are computed using $N=10$ generated hypotheses per object, leading to 140 generated hand-object interactions in total.

\subsubsection{Realism and Coverage of Human Grasps}

\begin{table}[h]
\centering
\caption{Grasp distribution analysis comparing generated grasps to ContactPose. Distances are symmetric surface distances (in mm). 
}
\label{tab:appendix_grasp_distribution}
\resizebox{0.7\columnwidth}{!}{
\begin{tabular}{llllll}
\toprule
\scriptsize Object & \scriptsize H. Var. & \scriptsize  Gen$\rightarrow$CP $\downarrow$ &  \scriptsize  CP$\rightarrow$Gen $\downarrow$ & \scriptsize  $D_{gen} \uparrow$ & \scriptsize  H. Div.\\
\midrule
\scriptsize Mug           & 6.4 & 8.3 & 11.3 & 18.6 &15.1 \\
\scriptsize Pan           & 6.0 & 11.6 & 12.5 & 15.8 &10.5\\ 
\scriptsize Apple         & 10.6 & 14.3 & 16.2 & 26.1 & 28.0\\
\scriptsize Hammer        &  6.0 & 8.7 & 10.2 & 11.9 &10.4\\
\scriptsize Knife         & 6.0 & 8.8 & 9.2 & 15.4 &10.2\\
\scriptsize Bottle        & 15.1 & 12.4 & 13.5 & 30.8 &24.3\\
\scriptsize Cellphone     &  6.5 & 9.1 & 9.2 & 30.4 &9.5\\ 
\scriptsize Cup           & 7.6 & 12.8 & 14.7 & 27.0 &31.1\\
\scriptsize Flashlight    & 5.5 & 11.4 & 11.8 & 16.6 &8.7\\
\scriptsize Scissors      & 6.5 & 9.0 & 10.7 & 27.3 &18.3\\
\scriptsize Stapler       & 7.1 & 12.4 & 12.8 & 24.7 &14.4\\
\scriptsize Toothbrush    & 6.4  & 11.5 & 10.3 & 26.4 &12.9\\
\scriptsize Teapot        & 6.7 & 11.8 & 10.6 & 17.6 &12.1\\
\scriptsize Wine Glass    & 10.5 & 9.1 & 14.5 & 31.6 &40.0\\
\bottomrule
\end{tabular}
}
\end{table}

To broaden our evaluation of the human grasp generation of GENESIS-Handover, we include two additional metrics, H. Var. and $D_{\text{gen}}$, as well as the full list of ContactPose~\cite{contactpose} objects that have a specific use case. For the ease of the reader, the explanations of the metrics already presented in Table \ref{tab:grasp_distribution} in Section \ref{sec:contactdb-analysis} are repeated in this section.

We compute symmetric nearest-neighbor distances between generated and ContactPose grasps. For each generated grasp, the generated-to-ContactPose ($Gen \rightarrow CP$) distance measures the distance to the closest human grasp and thus reflects the realism of the grasp. Conversely, the ContactPose-to-generated ($CP \rightarrow Gen$) distance measures, for each human grasp, the distance to the closest generated grasp and captures how well the generated set covers the diversity of human grasp configurations. Both metrics are reported as median distances (in mm) over their respective source sets.
Low $Gen\rightarrow CP$ distances indicate high realism, while low $CP \rightarrow Gen$ distances indicate good coverage of the human grasp distribution. 

To characterize the human grasps, we compute two statistics within the ContactPose dataset. We distinguish between variability and diversity. 
We define variability as the local consistency between grasps, measured by the median nearest-neighbor distance ($CP \rightarrow CP$), denoted as H. Var. 
In contrast, we define diversity as the global spread of grasp configurations, measured by the median pairwise distance ($D_{cp}$), denoted as H. 
Together, these metrics characterize the local and global structure of human grasping behavior: variability reflects the local consistency between neighboring grasps, while diversity captures the global spread and the presence of multiple grasp modes.

To assess the ability of the generative model to capture a range of grasp configurations, we compute diversity as the median pairwise symmetric surface distance between grasps. Given a set of grasps $\{g_i\}_{i=1}^M$, we define:
\[
D = \mathrm{median}_{i<j} \; d(g_i, g_j),
\]
where $d(\cdot,\cdot)$ denotes the symmetric surface distance between two grasps. We apply this definition both to the generated grasp set ($D_{\text{gen}}$) and to the ContactPose grasps ($D_{cp}$), enabling a direct comparison of generated and human grasp diversity.
Higher values indicate a broader exploration of grasp configurations.



Comparison of the grasp statistics presented in Table~\ref{tab:appendix_grasp_distribution} shows that coverage must be interpreted relative to both variability and diversity. Objects with low variability (H. Var.) and low diversity (H. Div.) (e.g., pan, flashlight, cellphone) exhibit highly consistent grasping behavior across users, while objects with higher diversity (e.g., mug, bottle, wine glass) exhibit multiple distinct grasp strategies.
Finally, comparing $D_{gen}$ to $D_{cp}$ shows that the method produces a comparable level of diversity for several objects, supporting its ability to capture diverse grasp configurations.



\subsubsection{Baseline Comparison}
\label{ap:baseline}
To contextualize the generated grasp hypotheses, we compare the predicted contact regions against AFT-Handover \cite{tulburehri26}, which represents a baseline for task-oriented grasp prediction.
AFT-Handover predicts task-specific graspable regions as surface area, modeling the average human contact distribution. In contrast, GENESIS-Handover generates individual, instance-specific grasp hypotheses.

We evaluate contact predictions using precision and recall with respect to ContactPose, measuring how accurately predicted contact regions align with human demonstrations and how much of the human contact area is covered. In addition, we report Pearson correlation to assess similarity to the aggregated human contact distribution and the active contact area to quantify spatial extent.

As shown in Table~\ref{tab:contact_aggregate}, AFT-Handover achieves higher recall and larger active areas, indicating broader but more diffuse contact predictions. In contrast, GENESIS-Handover produces more localized contact regions with higher precision, reflecting that individual generated grasps correspond to specific grasp configurations rather than covering the full contact region. At the distribution level, Pearson correlation with ContactPose is generally higher for AFT-Handover, as it explicitly approximates the averaged human contact distribution. The proposed method, however, often shows lower correlation, particularly for objects with high grasp diversity (e.g., cup or wine glass). This is expected, as the generated hypotheses capture distinct grasp modes rather than approximating their mean.

Overall, the results highlight the fundamental methodological difference between the two: AFT-Handover predicts a single, averaged contact distribution, while GENESIS-Handover represents multiple plausible grasp configurations. As a result, our approach provides more specific, task-consistent grasp hypotheses.

\begin{table*}[t]
\centering
\caption{Comparison to AFT-Handover~\cite{tulburehri26} on aggregate contact prediction. We report precision, recall, Pearson correlation with ContactPose, and active contact area.}
\label{tab:contact_aggregate}
\resizebox{0.9\textwidth}{!}{
\begin{tabular}{l|cccc|cccc}
\toprule
 & \multicolumn{4}{c|}{\scriptsize \textbf{GENESIS-Handover (aggregate)}} 
 & \multicolumn{4}{c}{\scriptsize  \textbf{AFT-Handover (aggregate)}} \\
\scriptsize \textbf{Object} 
& \scriptsize Prec. $\uparrow$ & \scriptsize Rec. $\uparrow$ & \scriptsize Corr. $\uparrow$ & \scriptsize Area $\downarrow$
& \scriptsize Prec. $\uparrow$ & \scriptsize Rec. $\uparrow$ & \scriptsize Corr. $\uparrow$ & \scriptsize Area $\downarrow$ \\
\midrule
\scriptsize Mug         & 0.96 & 0.08 & 0.62 & 0.05 & 0.89 & 0.14 & 0.59 & 0.10 \\
\scriptsize Hammer      & 1.00 & 0.23 & 0.57 & 0.22 & 1.00 & 0.32 & 0.91 & 0.31 \\
\scriptsize Pan         & 1.00 & 0.04 & 0.84 & 0.04 & 1.00 & 0.07 & 0.98 & 0.08 \\
\scriptsize Knife       & 1.00 & 0.40 & 0.81 & 0.39 & 1.00 & 0.38 & 0.86 & 0.38 \\
\scriptsize Bottle      & 1.00 & 0.40 & 0.44 & 0.33 & 0.98 & 0.47 & 0.67 & 0.40 \\
\scriptsize Scissors    & 1.00 & 0.38 & 0.49 & 0.38 & 1.00 & 0.33 & 0.73 & 0.33 \\
\scriptsize Cup         & 0.56 & 0.29 & -0.05 & 0.37 & 0.64 & 0.38 & 0.02 & 0.44 \\
\scriptsize Apple       & 1.00 & 0.60 & 0.11 & 0.59 & 1.00 & 0.52 & 0.37 & 0.51 \\
\scriptsize Flashlight  & 1.00 & 0.63 & 0.75 & 0.56 & 1.00 & 0.36 & 0.68 & 0.32 \\
\scriptsize Stapler     & 0.95 & 0.31 & 0.38 & 0.26 & 0.84 & 0.37 & 0.06 & 0.37 \\
\scriptsize Toothbrush  & 1.00 & 0.70 & 0.86 & 0.69 & 1.00 & 0.33 & 0.75 & 0.31 \\
\scriptsize Teapot      & 0.69 & 0.06 & 0.35 & 0.04 & 0.62 & 0.19 & 0.35 & 0.13 \\
\scriptsize Wineglass   & 0.78 & 0.09 & 0.27 & 0.08 & 0.59 & 0.60 & -0.05 & 0.65 \\
\scriptsize Cellphone       & 1.00 & 0.39 & 0.55 & 0.38 & 1.00 & 0.45 & 0.12 & 0.44 \\
\bottomrule
\end{tabular}
}
\end{table*}


\begin{figure}[h]
    \centering
     \includegraphics[width=1.0\linewidth]{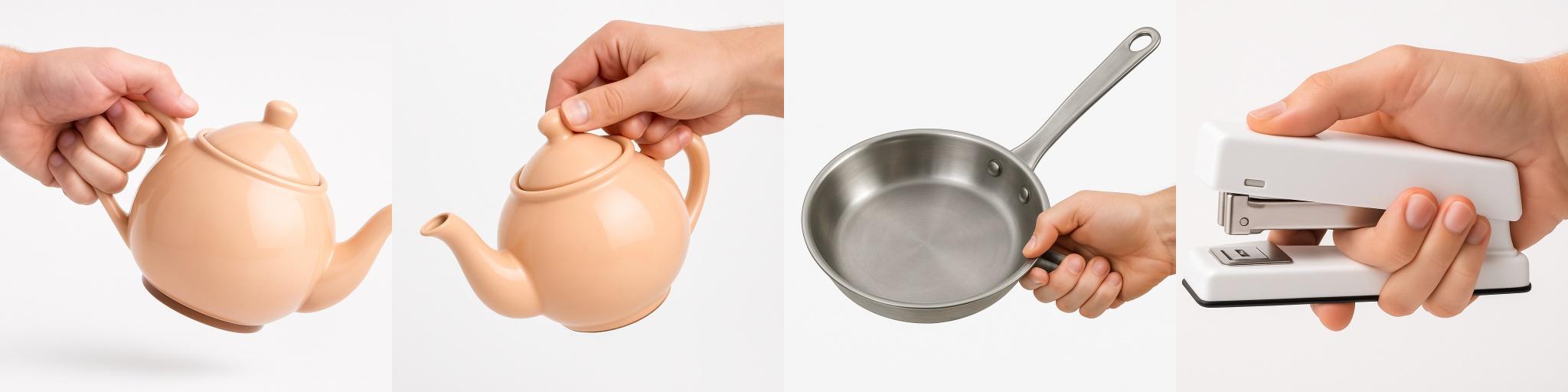}
     \caption{Generated image examples. 1 (left head instead of right) and 4 (infeasible grasp) are corrected by WiLoR, as the right is matched correctly. Images 2 and 3 lead to failures as the grasps are wrong.}
     \label{fig:ig-weird-grasps}
\end{figure}

\subsubsection{Failure Analysis}
A failure analysis reveals that 16 out of 140 generated hand-object interactions ($11\%$) are visually infeasible and would likely lead to unsuccessful handovers. The majority of these failures (12/16) originate from ICP misalignment, primarily due to insufficient geometric features, leading to ambiguous object orientations (e.g., 180° rotations for objects such as cellphone, bottle, and stapler). Additionally, image-generation artifacts occur in 12 out of 140 cases ($8.6\%$) with examples shown in Fig.~\ref{fig:ig-weird-grasps}. Those are typically minor (e.g., incorrect right vs. left hand or slight shape deviations) and rarely deviate from the distribution of human grasps observed in ContactPose \cite{contactpose}. Only 4 of these lead to errors in the relative hand-object configuration that fall outside the distribution. 


Overall, these results indicate that the generated hypothesis set captures realistic and diverse grasp configurations and provides a meaningful basis for hypothesis selection. At the same time, the main limitations arise from geometric alignment under ambiguity and ICP convergence, not from the quality of the generated grasp hypotheses themselves.
However, this analysis does not capture how these factors affect interaction in real-world robot-to-human handovers.



\subsection{Auxiliary User Study Evaluations}

In addition to the core findings of the user study presented in Section \ref{sec:user-study}, we present a more detailed analysis, including object-level user preferences and failure cases.

\subsubsection{System-Level Evaluation}
\begin{figure}[h]
    \centering
\includegraphics[width=1.0\linewidth]{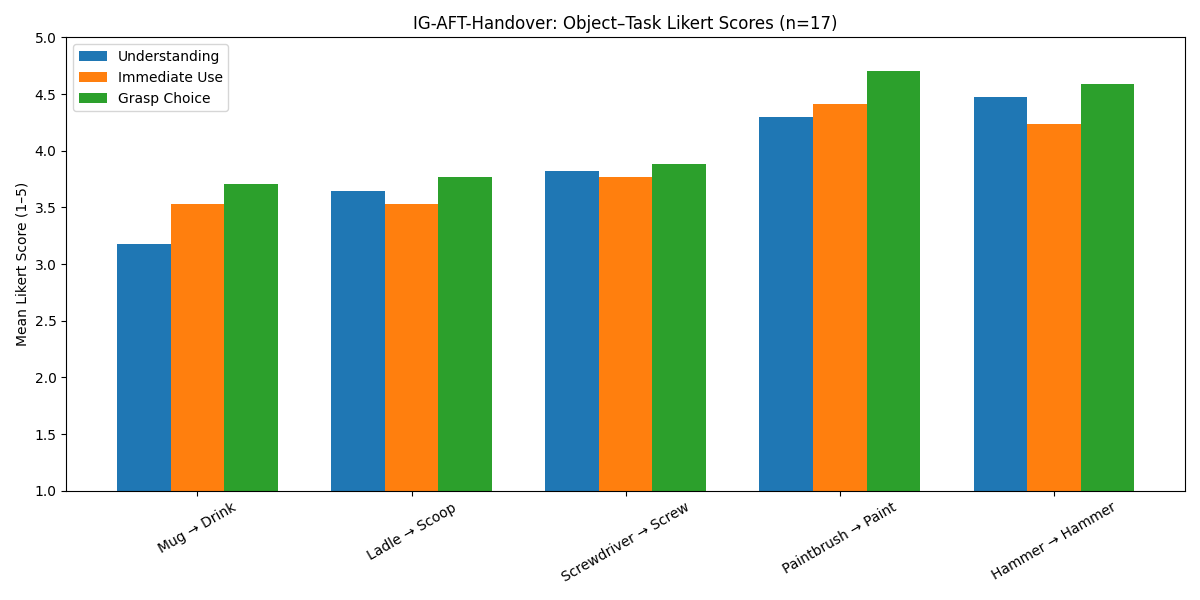}
    \vspace{-12pt}
    \caption{Mean Likert-scale rating (with 1 being strongly disagree and 5 strongly agree) for different object-task pairs and evaluation criteria.}
    \label{fig:ig-likert}
\end{figure}

We analyze GENESIS-Handover's performance across object-task pairs based on user ratings of task understanding, immediate usability, and grasp suitability. 
As summarized in Fig.~\ref{fig:ig-likert}, tool-based tasks such as paintbrush-paint and hammer-hammer receive the highest ratings across all dimensions, with paintbrush achieving the highest usability ($4.41\pm0.94$) and grasp suitability ($4.71\pm0.59$). Tasks such as ladle-scoop and screwdriver-screw yield intermediate ratings, reflecting moderate alignment with intended use.
Across these four object-task pairs, standard deviations are moderate, indicating consistent but not uniform subjective perception.
In contrast, the mug-drink task exhibits lower ratings (understanding: 3.18$\pm$1.24; usability: 3.53$\pm$1.63) and higher variance. While handle accessibility is ensured, the system does not enforce upright orientation, which appears to be an implicit user expectation for container-based tasks. This suggests that beyond affordance alignment, task-oriented handovers may require reasoning about physical constraints.

Overall, the results demonstrate that GENESIS-Handover performs well for tasks where the intended use strongly constrains object orientation, while performance decreases for tasks with additional implicit physical requirements.

\subsubsection{Failure Analysis}
The following analysis is based exclusively on handovers performed with GENESIS-Handover during the user study and examines performance in a real-world deployment setting. Beyond the 60 handovers collected in the main study, we include an additional 25 handovers conducted under identical conditions. As these participants were not exposed to the baseline method, their data are considered only for system-level evaluation and are excluded from direct comparisons between methods.
Across all 85 GENESIS-Handover handovers, 7 resulted in failures as defined in Sec.~\ref{sec:eval-metrics}. These failures consisted of 4 grasp mismatches and 3 significant pose errors.
We attribute them to the following components of the pipeline: one failure originated from AFT-Handover, 2 from erroneous live hand pose estimation, 2 from motion planning errors, and 2 from ICP misalignment.


In addition to these failures, a larger number of intermediate errors were observed during the user study that did not lead to failed handovers, but to adjustments after the handover. These include 4 cases of incorrect live hand pose estimation, 6 motion planning issues, and both major (3 cases) and minor (5 cases) ICP misalignments. Generation-related artifacts (inaccurate object geometry, incorrect handedness, weird grasps) were observed in 12 cases but only affected the outcome in 3. This indicates that the system exhibits robustness to intermediate errors in real-world interaction. 

\begin{table}[h]
\centering
\caption{Hypothesis diversity measured as the number of clusters per handover. Clusters are obtained via hierarchical clustering (average linkage) with a fixed distance threshold. Hand clusters are computed from hand-configuration distances, while combined clusters additionally incorporate normalized RGB hand-object relative position. Values report mean cluster counts over handovers.}
\label{tab:hypothesis_clusters}
\vspace{5pt}
\resizebox{0.9\linewidth}{!}{
\begin{tabular}{lcccccccc}
\hline
\multirow{2}{*}{\textbf{Object}} 
& \multicolumn{4}{c}{\textbf{Hand Config.}} 
& \multicolumn{4}{c}{\textbf{Hand Config. + RGB}} \\
\cline{2-5} \cline{6-9}
 & Mean & Std & Min & Max & Mean & Std & Min & Max \\
\hline
Hammer      & 1.35 & 0.49 & 1 & 2 & 2.81 & 0.54 & 2 & 4 \\
Ladle       & 1.67 & 0.62 & 1 & 3 & 2.80 & 0.41 & 2 & 3 \\
Mug         & 2.63 & 0.50 & 2 & 3 & 2.19 & 0.40 & 2 & 4\\
Paintbrush  & 1.13 & 0.35 & 1 & 2 & 2.63 & 0.59 & 2 & 3 \\
Screwdriver & 1.94 & 0.25 & 2 & 2 & 2.88 & 0.62 & 2 & 4 \\
\hline
\end{tabular}
}
\end{table}

\subsection{Hypothesis Structure and Ambiguity}
\label{ap:hyp_structure}
In the following, we analyze the structure of the hypothesis space and the resulting ambiguity at handover onset.
We first examine the organization of the hypothesis set. We compute pairwise distances between hypotheses in the hand-configuration space. Hierarchical clustering reveals that the hypotheses form a small number of grasp modes, typically one to two per object, as shown in Table~\ref{tab:hypothesis_clusters}.
When incorporating object-relative interaction, which captures the relative placement of the hand on the object, the number of clusters increases. This shows that similar hand configurations can occur at different locations on the object. These clusters often correspond to separate functionally meaningful grasp strategies.
For instance, a power grasp applied at different points along a hammer handle leads to similar hand poses but distinct interaction regions. Importantly, these interaction regions are often functionally meaningful, as they correspond to different ways of using the object, such as applying a higher hammering force versus enabling more precise manipulation.

\begin{table}[t]
\centering
\caption{Ambiguity statistics per object. We report the fraction of ambiguous handovers, the average number of competing hypotheses within the ambiguity threshold, and whether ambiguity occurs within a single cluster or across clusters.}
\label{tab:ambiguity}
\vspace{5pt}
\resizebox{0.8\linewidth}{!}{

\begin{tabular}{lcccc}
\hline
\textbf{Object} & \textbf{Ambig. (\%)} & \textbf{\#Hyp.} & \textbf{\#Clusters} & \textbf{Intra (\%)} \\
\hline
Hammer      & 35.3 & 1.88 & 1.06 & 94.0 \\
Ladle       & 46.7 & 1.87 & 1.00 & 100.0 \\
Mug         & 18.8 & 1.19 & 1.00 & 100.0 \\
Paintbrush  & 81.3 & 2.25 & 1.00 & 100.0 \\
Screwdriver & 56.3 & 2.00 & 1.06 & 94.0 \\
Average     & 47.5 & 1.84 & 1.02 & 98.0 \\
\hline
\end{tabular}
}
\end{table}

This structured hypothesis space implies that the selection problem should be solved not by searching over a large set of candidates, but by disambiguating between a small number of competing hypotheses. To quantify this ambiguity, we analyze the relative gap between the best and second-best selected hypothesis, using a $10\%$ threshold to identify near-ties. As shown in Table~\ref{tab:ambiguity}, a substantial fraction of handovers ($47.5\%$) exhibit ambiguity, indicating that at least two hypotheses are similarly consistent with the initial hand observation.
Importantly, this ambiguity is typically localized: in most cases, only one or two hypotheses fall within the ambiguity threshold, and they belong to the same cluster. This indicates that the correct coarse grasp mode is often identified, while ambiguity arises from fine-grained variations within that mode. As a result, even imperfect selection can still lead to functionally usable handovers.



\begin{figure}[h!]
    \centering
    \includegraphics[width=0.85\linewidth]{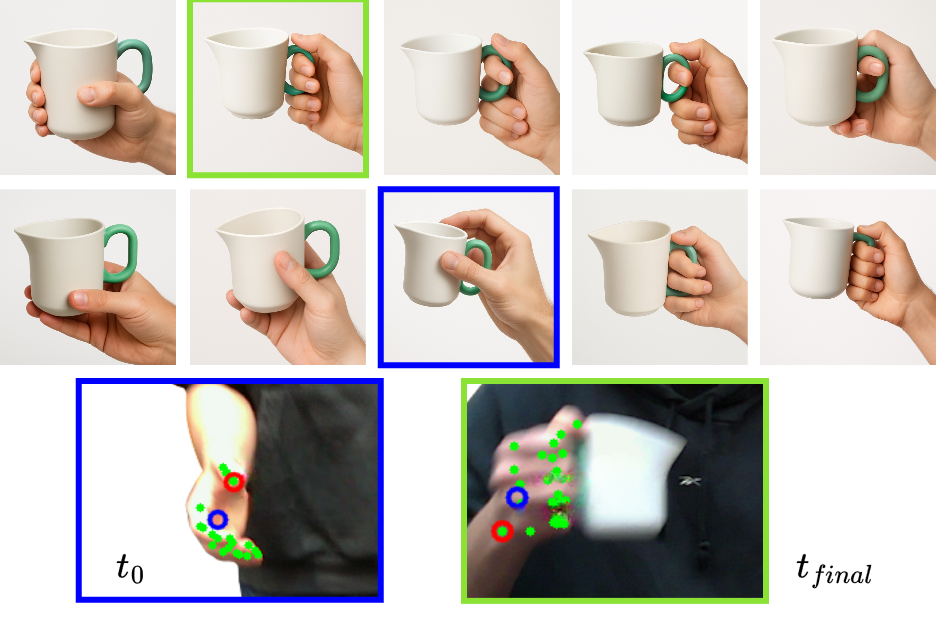}
    \vspace{-8pt}
    \caption{Failure case example on a mug. Among multiple plausible grasp hypotheses, the oracle grasp (green) aligns with the human’s final grasp, while the selected hypothesis (blue) corresponds to another strategy. This highlights the difficulty of ranking valid grasps from initial observations under ambiguity.}
    \label{fig:mug_failure}
\end{figure}
\begin{figure}[h!]
    \centering
    \includegraphics[width=0.85\linewidth]{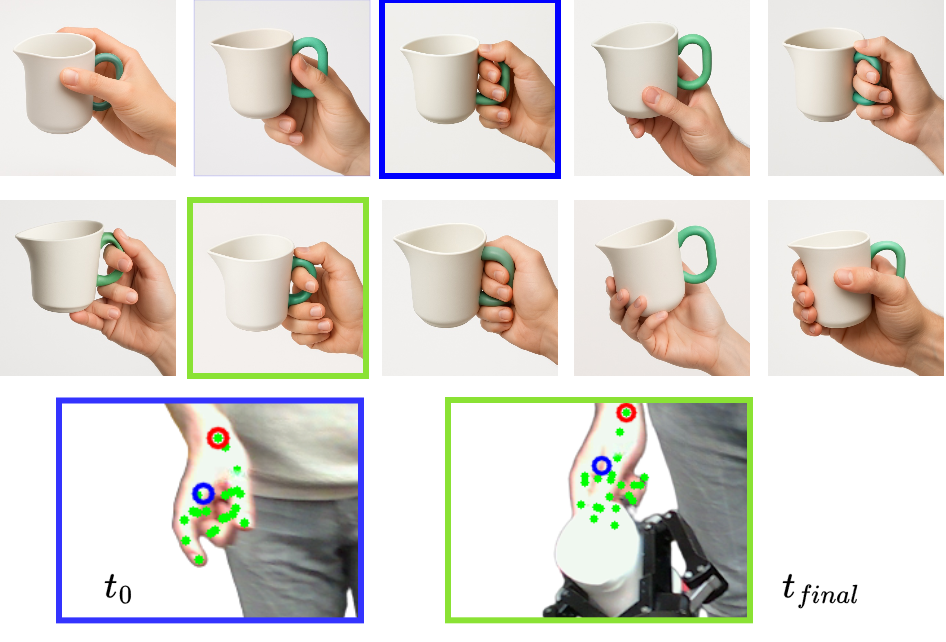}
    \vspace{-8pt}
    \caption{Near-miss case for a mug. The selected hypothesis (blue) closely matches the oracle (green) and the final human grasp, with only minor differences. The initial hand pose already reflects the final grasp configuration, reducing ambiguity and enabling accurate selection.}
    \label{fig:mug_borderline}
\end{figure}

For the mug, selection performance was noticeably weaker than for other objects, suggesting that in this case, ambiguity might be a limiting factor. However, Table~\ref{tab:ambiguity} indicates a low ambiguity rate ($18.8\%$) within a cluster. This appears to contradict previous results. While the closest hypotheses are mutually consistent, they do not necessarily correspond to the correct grasp strategy. Often, the top-ranked hypotheses all belong to the same (incorrect) cluster, while the oracle lies in a different cluster and is not among the nearest candidates to the initial hand configuration. 
This behavior is illustrated in Fig.~\ref{fig:mug_failure}, where both the selected and second-best hypotheses at time $t_0$ correspond to a power grasp, while the oracle is a handle grasp from a different cluster. The initial selection is therefore locally consistent given the observed hand, but globally incorrect with respect to the intended grasp.

In contrast, Fig.~\ref{fig:mug_borderline} shows a case where the initial hand pose more clearly reflects the final grasp, resulting in a selection that closely matches the oracle. These examples illustrate that the quality of hypothesis selection depends on the discriminative power of the early hand observation with respect to the intended grasp. 

\subsection{Human Handover Behavior Analysis}
\label{ap:handover_behav}
We analyze the temporal evolution of hand configurations during the approach phase by measuring joint-level distances relative to the initial hand pose and identifying discrete reconfiguration events. Across all handovers from the user study with GENESIS-Handover, hand configurations remain highly stable, with a median deviation of $0.006$, and $75\%$ of handovers stay below the significant-change threshold of $0.027$. The final pre-grasp configuration also remains close to the initial observation (median $0.012$), indicating that hand configurations remain largely consistent throughout the approach phase.

Next, we analyze reconfiguration events, defined as observable changes ($0.027$) in hand articulation. Such events occur in $27.1\%$ of handovers, but are not uniformly distributed: while some of them occur early during approach ($34\%$), most happen close to grasp execution ($66\%$). Importantly, the early changes do not consistently correspond to changes in grasp intent. Most users return to configurations similar to the initial pose after intermediate adjustments (in $62\%$ of early hand reconfiguration cases), indicating that these variations reflect local motion rather than different grasp strategies.
Furthermore, only $50\%$ of hand reconfiguration events are captured by the onboard perception system, primarily due to occlusions during late-stage interaction. This further limits the practical observability of such changes.

These observations indicate that early hand configurations are generally stable and provide a consistent signal throughout the approach phase. Although variations in hand pose do occur, they are infrequent and typically arise late in the interaction, suggesting that continuous re-selection is unlikely to provide substantial benefit in most cases. 

\section{Supplementary Information}
\subsection{User Study Ethics Approval}
The proposal detailing the user study presented in this work was reviewed and approved by the ETH Zurich Ethics Commission. The official proposal title and reference number were as follows: \emph{Project 26 ETHICS-031: Human-Legged Robot Handover System Evaluation}.

\subsection{Hypothesis Generation Prompting}
For completeness, we include the code used to generate the text prompts for both the grasp-style descriptions and the hypothesis images:

\begin{lstlisting}[language=Python]

def _generate_grasp_styles(self, object_str, task_str):
    n = self.num_images
    system_msg = (
        "You are a robotics and human-factors expert specialising in grasping. "
        "Your output must be a numbered list with exactly {n} items and nothing else. "
        "Each item is one concise sentence (<=20 words) describing HOW a human RIGHT hand "
        "grasps the object to do the specified task - covering grip type, finger placement, and contact region. "
        "First emphasis is on realistic grasps, suitable for the specified task, while being as diverse as possible:"
        "different finger configurations and contact region."
    ).format(n=n)
    user_msg = (
        f"Object: {object_str}\nTask: {task_str}\n"
        f"Generate {n} maximally diverse grasp styles."
    )
    response = self.openai_client.chat.completions.create(
        model="gpt-5.3",
        messages=[
            {"role": "system", "content": system_msg},
            {"role": "user",   "content": user_msg},
        ],
        temperature=1.0,
        max_completion_tokens=400,
    )
    raw = response.choices[0].message.content.strip()
    styles = []
    for line in raw.splitlines():
        line = line.strip()
        if not line:
            continue
        cleaned = line.lstrip("0123456789.-) ").strip()
        if cleaned:
            styles.append(cleaned)
    rospy.loginfo("Generated %d grasp styles:", len(styles))
    for i, s in enumerate(styles):
        rospy.loginfo("  [%d] %s", i, s)
    return styles[:n]
    

def _generate_images(self, ref_bgr, object_str, task_str, reason):
    rgb = cv2.cvtColor(ref_bgr, cv2.COLOR_BGR2RGB)
    pil_ref = PILImage.fromarray(rgb)
    buf = io.BytesIO()
    pil_ref.save(buf, format="PNG")
    png_bytes = buf.getvalue()

    grasp_styles = self._generate_grasp_styles(object_str, task_str)

    ros_images = []
    for i, grasp_style in enumerate(grasp_styles):
        prompt = (
            f"Generate a studio photo on a plain white background of a RIGHT hand holding a {object_str} "
            f"to {task_str}. {grasp_style}."
            f"Pay attention to the {object_str} details and make sure the generated {object_str} looks EXACTLY "
            f"like the one in the given image. Ensure that all the subpart geometries match! "
            f"Constraint1: Do not show any environment or additional objects, solely the human hand grasping "
            f"the {object_str}! "
            f"Constraint2: The hand must be in a natural and realistic pose for the specified grasp style, and the {object_str} must be oriented appropriately for the task. "
        )

        response = self.openai_client.images.edit(
            model="gpt-image-1",
            image=("reference.png", png_bytes, "image/png"),
            prompt=prompt,
            size="1024x1024",
            n=1,
        )

        b64_img = response.data[0].b64_json
        gen_pil = PILImage.open(
            io.BytesIO(base64.b64decode(b64_img))
        ).convert("RGB")

        gen_np = np.array(gen_pil)
        gen_np = cv2.resize(gen_np, (512, 512), interpolation=cv2.INTER_AREA)
        gen_bgr = cv2.cvtColor(gen_np, cv2.COLOR_RGB2BGR)

        gen_ros = self.bridge.cv2_to_imgmsg(gen_bgr, encoding="bgr8")
        gen_ros.header.stamp = rospy.Time.now()

        rospy.loginfo("[%s] Image %d/%d generated (%s)", reason, i + 1, len(grasp_styles), grasp_style[:40])
        ros_images.append(gen_ros)

    return ros_images

\end{lstlisting}

\end{document}